\documentclass[10pt,twocolumn,letterpaper]{article}
\usepackage[dvipsnames]{xcolor}
\usepackage[normalem]{ulem}
\usepackage{multirow}
\usepackage{wacv}
\usepackage{times}
\usepackage{epsfig}
\usepackage{graphicx}
\usepackage{amsmath}
\usepackage{amssymb}
\usepackage{booktabs}

\usepackage[accsupp]{axessibility}  

\def\wacvPaperID{97} 

\wacvapplicationstrack 

\wacvfinalcopy 

\ifwacvfinal
\usepackage[breaklinks=true,bookmarks=false,colorlinks=true,urlcolor=blue]{hyperref}  

\else
\usepackage[pagebackref=true,breaklinks=true,colorlinks,bookmarks=false]{hyperref}
\fi

\begin{document}

\title{PHG-Net: Persistent Homology Guided Medical Image Classification}

\author{Yaopeng Peng\\
University of Notre Dame\\
{\tt\small ypeng4@nd.edu}
\and
Hongxiao Wang\\
University of Notre Dame\\
{\tt\small hwang21@nd.edu}
\and
Milan Sonka\\
University of Iowa \\
{\tt\small milan-sonka@uiowa.edu}
\and
Danny Z. Chen\thanks{This research was supported in part by NIH NIBIB Grant R01-EB004640.}\\
University of Notre Dame \\
{\tt\small dchen@nd.edu}
}

\maketitle
\thispagestyle{empty}

\begin{abstract}
Modern deep neural networks have achieved great successes in medical image analysis. However, the features captured by convolutional neural networks (CNNs) or Transformers tend to be optimized for pixel intensities and neglect key anatomical structures such as connected components and loops. In this paper, we propose a persistent homology guided approach (PHG-Net) that explores topological features of objects for medical image classification. For an input image, we first compute its cubical persistence diagram and extract topological features into a vector representation using a small neural network (called the PH module). The extracted topological features are then incorporated into the feature map generated by CNN or Transformer for feature fusion. The PH module is lightweight and capable of integrating topological features into any CNN or Transformer architectures in an end-to-end fashion. We evaluate our PHG-Net on three public datasets and demonstrate its considerable improvements on the target classification tasks over state-of-the-art methods.
\end{abstract}

\section{Introduction}
Deep neural networks (DNNs) are capable of learning useful image features based on their potent representations, and are widely used in medical image analysis. From AlexNet~\cite{alexnet}, VGG~\cite{vggnet}, to DenseNet~\cite{googlnet,resnet,densenet}, many convolutional neural network (CNN) architectures have been proposed for rich feature representations. Due to their nature, CNNs primarily focus on capturing local features. Vision Transformers~\cite{dosovitskiy2020image, liu2022swin} have been proposed to procure global dependencies and long-range relationships by leveraging self-attention mechanism, which allows models to attend to different patches and learn their dependencies. However, these models tend to neglect key global and robust anatomical structures (e.g., topological structures), such as connected components, loops, and voids. Medical images commonly contain tissues, organs, and lesions as connected components with specific patterns (e.g., loops and voids), but such structures and typologies are often overlooked by deep learning (DL) models.
 
In recent years, topological data analysis (TDA)~\cite{topo_intro} has been applied as a powerful methodology to analyze data in chemistry~\cite{chemistry}, medicine~\cite{medicine}, biology~\cite{biology}, and other fields. Persistent homology (PH) is a most widely-used method of TDA. It tracks topological changes of object dynamics during the filtration process, where a lifespan is associated with these changes in the form of entity birth or death. The collection of such birth-death time pairs forms a persistence diagram (PD). But, effectively utilizing persistence diagrams in machine learning is not straightforward due to the multi-set nature of the persistent homology building process.

A technique was proposed to input topological signatures into DNNs for 2D object shape and social network classification~\cite{hofer2017deep}. Additionally, a readout operation to aggregate node features into a graph representation was introduced~\cite{hofer2020graph}. Persistent landscapes were presented~\cite{pers_land} as a method to summarize topological data into vectors to use in machine learning. In~\cite{pers_image}, persistent images were proposed as a stable representation that converts a persistence diagram into a finite-dimensional vector representation. A framework in \cite{perslay} first encoded a persistence diagram into a vector and then learned the vectorization using a neural network. Another method was developed in \cite{pllay} to first encode a persistence diagram into a list of persistent landscapes and then vectorize them for use in DL models

\begin{figure*}
\centering
\includegraphics[width=1.9\columnwidth]{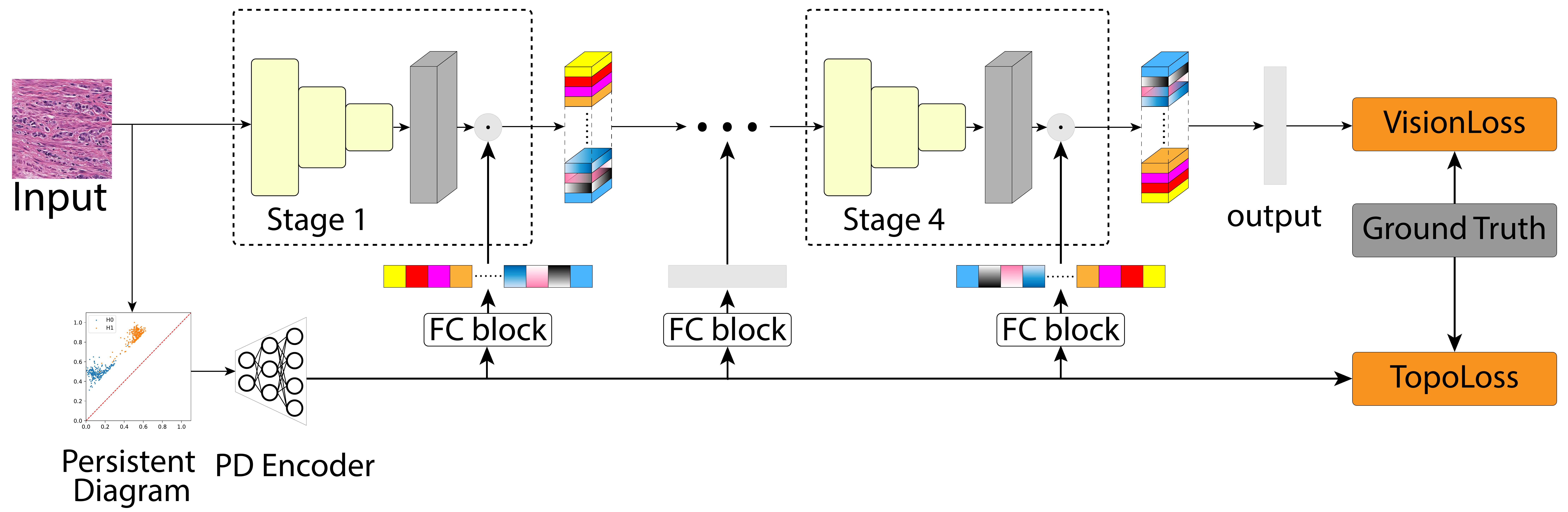}
\caption{Illustrating the pipeline of our proposed PHG-Net. The details of the PD encoder are shown in Fig.~\ref{ph_res}(a). A fully connected block (FC block) is a two-layer MLP shown in Fig.~\ref{ph_res}(b). The \textcolor{blue}{blue} points in a persistence diagram denote the 0-dimensional persistent homology (H0), and the \textcolor{BurntOrange}{orange} points denote the 1-dimensional persistent homology (H1). VisionLoss and TopoLoss represent the vision loss and topological loss, respectively. $\odot$ denotes matrix multiplication. We use a CNN backbone for illustration. 
 }
\label{pha}
\end{figure*}

Medical images often exhibit intricate marked topological structures/patterns of clinical targets, thereby rendering persistent homology (PH) a promising computational technique that provides supplementary topological information and insights alongside pixel-oriented CNNs~\cite{topo_intro}. More specifically, cubical persistence leverages image intensity values as filtration functions in its analytical process, which serves as a persistent homology tool that has been proven to be highly useful for medical image analysis.
For example, in \cite{toporesnet}, it computed persistent curves and statistics, and these features were integrated into ResNet for skin lesion classification. In \cite{prostate_cancer}, the persistence of each region of interest (e.g., \textit{birth\_time}-\textit{death\_time}) was ranked, and the images were clustered into sub-architectural groups. In \cite{qaiser2019fast}, topological features were combined with features extracted from the last CNN layer for tumor segmentation in histology slides. In \cite{topotxr}, it first masked out the critical positions of breast MRI images using persistent homology and classified the masked images. In \cite{du2022distilling}, a persistence diagram was first encoded into Betti curves and then integrated into a CNN network. In \cite{TopAtt}, it built topological-attention of consecutive slices for 3D anisotropic image segmentation. These methods often involved encoding a persistence diagram into a vector using mathematical tools, which is usually a non-trivial process. Another aspect that could be improved is that they concatenated topological features with the last CNN layer, which incorporates another type of (CNN) features, but lacks interactions between the topological features and multi-scale CNN features. Furthermore, the topological features are not involved when updating the gradients of the CNN or Transformer weights.


To address these issues, we propose a persistent homology-guided approach (PHG-Net) for medical image classification. PHG-Net directly processes persistence diagrams with a neural network and leverages the extracted topological features in multiple layers of CNN or vision Transformer (not just the last layer) to refine vision features at multiple scales. We treat persistence diagrams as point clouds and process them with a network (called the PH module or topological branch) inspired by PointNet \cite{pointnet}. Code is available at \href{https://github.com/yaoppeng/TopoClassification}{https://github.com/yaoppeng/TopoClassification}

Our main contributions are three-fold:
\begin{enumerate}

\item Inspired by PointNet \cite{pointnet}, we develop a new approach to process persistence diagrams as point clouds rather than vectorized features, using a neural network. The motivation for our work is that the differences among topological structures of images can be represented by persistence diagrams and captured by a neural network, thus guiding the learning of CNNs or Transformers. This process is data-driven and learnable.
    	
\item Our PHG-Net is capable of incorporating topological features into any base vision models, including CNNs and Transformers.
     
\item  Our new PHG-Net approach achieves considerable improvements on three public datasets compared to state-of-the-art medical image classification methods.
\end{enumerate}

\section{Method}
We use a CNN backbone to illustrate our PHG-Net.
The overall pipeline of our PHG-Net approach is shown in Fig.~\ref{pha}. Given an image $I$, the persistence diagram (PD) of $I$ is first computed, and a PointNet-like neural network is then applied to encode the PD into a feature vector. The feature vector is integrated into feature maps generated by the CNN at each of its ConvBlocks for feature map refinement.

In this section, we first provide a brief review of cubical persistent homology (the readers may refer to~\cite{topo_intro} for further exposition). We then describe how we design a neural network encoder to directly convert a PD into a feature vector. Finally, we present the process of using persistent homology to guide the CNN for feature refinement in medical image classification.

\begin{figure*}[ht]
\centering

\includegraphics[width=1.7\columnwidth]{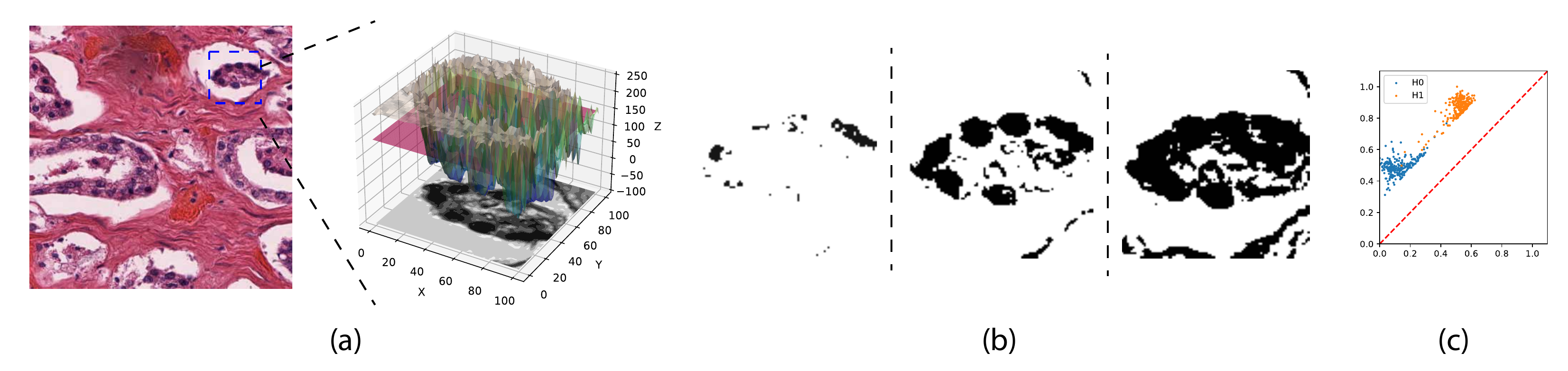}

\caption{Illustrating the process of sub-level filtration. (a) A 3D  plot of the sub-level set filtration of pixel intensities of a crop in an H\&E stained image. The corresponding hematoxylin channel of the crop is projected on the $xy$ plane. The $z$-axis is for threshold values. The red plane corresponds to one 2D slice of the 3D plot at threshold = 150. (b) Three thresholded binary images of the cropped image. As the threshold value increases or decreases, some connected components or loops are born or die. (c) The computed persistence diagram of the image crop (\textcolor{blue}{blue} points denote the 0-D persistent homology (H0), and \textcolor{BurntOrange}{orange} points denote the 1-D persistent homology (H1)). 
 }
\label{filtration}
\end{figure*}

\subsection{Cubical Persistent Homology}
Persistent homology (PH) is a powerful mathematical tool for analyzing topological properties of data. It is capable of identifying and quantifying important features of data that persist over a range of different spatial scales, thus providing insight into the underlying structures of the data. Fig.~\ref{filtration} illustrates the filtration process of PH. Given an image $I$ and a series of threshold values $(\tau_0, \tau_1, \ldots, \tau_h)$, $\tau_0 < \tau_1 < \cdots < \tau_h$, in each step $i$, we threshold $I$ as a sub-level image: $S_{\tau_i} = \{x: x \in I, f(x)\leq \tau_i\}$, where $f$: $I \rightarrow R$ is a function of pixel intensities. 
That is, for each $\tau_i$, a sub-image $S_{\tau_i}$ is generated, and its connected components, loops, and voids are recorded. The sequence of 
$\{S_{\tau_i} \ | \ i=0, 1,\ldots, h\}$, with $S_{\tau_0} \subseteq S_{\tau_1} \subseteq \cdots \subseteq S_{\tau_h}$, forms the filtration. Following the evolution of these sub-level images through the threshold sequence, the homology groups are induced as $\{H(S_{\tau_0}), H(S_{\tau_1}), \ldots, H(S_{\tau_h})\}$, where $H(S_{\tau_i})$ records the topological features of $S_{\tau_i}$ (e.g., connected component, loops, voids).  When a new topological structure appears or is ``born'' at threshold $\tau_i$ and disappears, ``dies'', or merges with another topological structure at threshold $\tau_j$, a tuple $(\tau_i, \tau_j)$ is recorded and plotted as a 2D point with $\tau_i$ on the $x$-axis and $\tau_j$ on the $y$-axis; $\tau_j - \tau_i$ denotes the lifespan (i.e., persistence) of that topological structure. A structure with a long persistence means that the differences between it and its surroundings are significant and tend to be more salient (e.g., the boundary region of an image).

\subsection{Persistence Diagram Encoder}
\label{sec-PD-encoder}

Since PD is a most commonly used descriptor for persistent homology, there are an increasing number of methods to map PDs into a vector representation for machine learning tasks. However, these methods often rely on a mathematical encoding step, which is usually non-trivial and specifically designed, and may result in loss of certain information of the PDs. To address this issue, we propose a data-driven method to encode PDs.

Specifically, we treat each data point in a PD as a 2D feature, denoted as $(f_{born}, f_{death})$. A straightforward method would be to aggregate all the data points and obtain a 2D vector, for example, through max pooling or average pooling. However, such an operation may lead to losing too much information of the PD and may not effectively represent the entire filtration process. Therefore, we first increase the dimension of each data point so that the points can interact with one another. Then, we aggregate the information of all the points into a vector representation.

\begin{figure*}[t!]
 \centering
 \includegraphics[width=2\columnwidth]{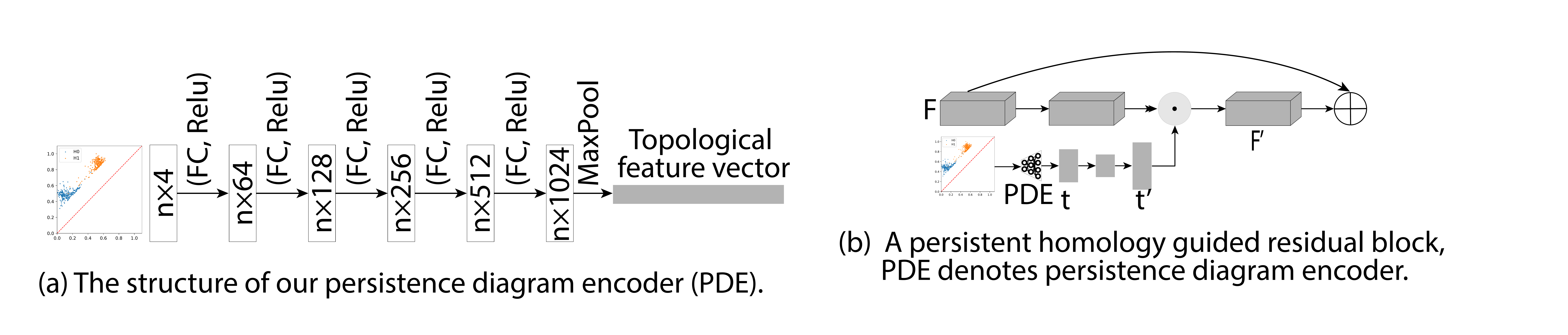}
	\caption{Illustrating (a) our persistence diagram encoder and (b) a persistent homology guided residual block (FC block). $\odot$ and $\oplus$ are matrix multiplication and summation, respectively. }
	\label{ph_res}
\end{figure*}

One key property of the points in a PD is permutation invariance, which means that the feature vector of the PD remains the same if some of the points exchange positions with one another. Note that this property is different from that of processing image data with a CNN, where exchanging pixels may result in different outputs. Based on this observation, we formulate the PD-encoding problem as:
\begin{equation}
f(p_1, p_2, \ldots, p_{n}) = g(m(p_1), m(p_2), \ldots, m(p_{n})),
\label{eq-PD-encoding}
\end{equation}
where $p_i$ denotes a data point in a PD, $f(p_0, p_1, \ldots, p_{n-1})$ represents the output feature vector, and $m(p_i)$ is a function that maps a data point into a higher dimension so that it contains interactions of the data points and the information of the entire PD. To account for the permutation invariance property, we use a multi-layer perceptron (MLP) instead of using convolution. In Eq.~(\ref{eq-PD-encoding}), $g$ is a function that aggregates information from all the points, such as through max pooling or average pooling. The entire vector learning process is carried out with a neural network, which is data-driven, and the vectorization process is supervised by the target task. Fig.~\ref{ph_res}(a) illustrates the details of the persistence diagram encoder that we design.

For persistent homology, let $H_0$ denote the connected components, $H_1$ denote the loops, and $H_2$ denote the voids in the 3D representation. However, a neural network taking points from these groups as input cannot differentiate which homology group that a point belongs to. Therefore, we add an additional one-hot vector to label each point. In other words, a point $(birth, death)$ in $H_i$ is represented as $(birth, death, 0, \ldots, 0, 1, 0, \ldots, 0)$, in which the only 1 is located in the $i$-th position of the 0-1 sequence.

\subsection{Persistent Homology Guidance (PHG)}

The persistent homology features capture global topological information of an image. Previous methods typically utilized these features by integrating them with features learned by CNNs or Transformers. For example, in \cite{toporesnet}, it concatenated PH features and CNN features just before the classification step. In \cite{du2022distilling}, PH features were treated as a teacher for adjusting CNN features. However, in concatenation based methods, gradient values of multiple branches during back-propagation are computed separately, and there are no direct interactions between the gradients of different branches. Moreover, concatenation just before the classification step lacks interactions between topological features and CNN features at multi-scale levels.
To tackle this issue, we propose integrating topological supervision with each CNN or Transformer block. Additionally, we add a fully connected block (FC block) to regulate the intervention of the topological branch in each CNN or Transformer block (see Fig.~\ref{ph_res}(b)). This also enables us to refine the feature maps in multi-scales using topological features.

Given an intermediate CNN feature map $F$, $F \in \mathbb{R}^{H \times W \times C}$, our PHG-Net integrates a topological feature vector $t$ into $F$, where $t \in \mathbb{R}^M$ is the output of the persistent homology encoder (see Eq.~(\ref{eq-PD-encoding})). To ensure flexibility and applicability to any feature map, we employ a gate mechanism in the process:
\begin{equation}
	t' = f(t, W) = \sigma(W_1(Relu(W_2(t)))),
\end{equation}
where $t'\in R^C$ is the processed topological feature vector that will be used to refine the feature map $F$, $t$ denotes the output of the PD encoder, $W_1\in R^{ \frac{C}{r}\times C}$ and $W_2\in R^{M\times \frac{C}{r}}$ are the parameters of two MLP layers, $\sigma$ is the sigmoid activation function, and $r$ is a parameter for balancing the trade-off between the performance and complexity. Afterwards, the topological guidance is formulated as:
\begin{equation}
	F' = F\otimes t',
\end{equation}
where $F'$ is the refined feature map and $\otimes$ is element-wise multiplication. During the refinement, the feature vector $t'\in R^C$ will first be broadcast along the spatial dimension of $F'$. Fig.~\ref{ph_res}(b) illustrates a persistent homology guided residual block. 

\subsection{Loss Function}
The whole model is optimized with the following loss ${\cal{L}}$:
\begin{equation}
    {\cal{L}} = {\cal{L}}_{V}(\tilde{y}_{v}, y) + \alpha {\cal{L}}_{Topo}(\tilde{y}_{topo}, y),
    \label{eq-final-loss}
\end{equation}
where ${\cal{L}}_{V}$ and ${\cal{L}}_{Topo}$ are the cross-entropy losses of the vision model and topological branches respectively; $\tilde {y}_{v}$, $\tilde{y}_{topo}$, and $y$ denote the vision branch output, topological branch output, and ground truth respectively, and $\alpha$ is a hyper-parameter for balancing the CNN and topological losses. See Fig.~\ref{pha} for more details.

\section{Experiments}
\subsection{Datasets}
We evaluate our proposed PHG-Net approach using the following three datasets.

\noindent
{\bf ISIC 2018:} A skin lesion dataset for predicting 7 classes of skin disease lesions. It consists of 10,015 training images and 193 validation images~\cite{codella2019skin}; the test set is 
unavailable. For fair comparison, we follow the dataset split strategy in~\cite{zhuang2018skin}, with which five-fold cross validation is conducted.

\noindent
{\bf Prostate Cancer Classification:} A dataset of hematoxylin and eosin (H\&E) stained prostate cancer images that consists of 77 whole slide images (WSIs) from 19 patients~\cite{prostate_cancer}. The WSIs are divided into regions of interest (RoIs) of $512\times 512$ pixels each. 5,182 RoIs are generated and three classes of prostate cancer are to be predicted. We also conduct five-fold cross validation.

\noindent
{\bf CBIS-DDSM:} A Curated Breast Imaging Subset of Digital Database for Screening Mammography (CBIS-DDSM). It contains 1,566 participants and 6,775 studies, which are categorized as benign or malignant~\cite{lee2017curated}. We use the train/test split given by the dataset provider, in which 20\% of the cases are for testing and the rest for training.

\subsection{Experimental Setup}
\begin{table*}[ht]

\centering

\caption{Experimental results on the ISIC 2018 dataset. 
Reported $p$-values reflect statistical significance of the improvement achieved by SwinV2-B + PHG over SwinV2-B, assessed by paired t-test; ``--'' is reported if SwinV2-B + PHG performance is lower than SwinV2-B for a specific evaluation metric.
}


\label{main_isic}

\begin{tabular}{l|c|c|c|c}
Method &  Accuracy & AUC & Sensitivity & Specificity \\
\hline
		
ResNet152~\cite{resnet} & 88.83$\pm$0.45 & 97.52$\pm$0.15 & 80.85$\pm$0.16 & 97.01$\pm$0.28 \\

ResNet152~\cite{resnet} + PHG & 90.03$\pm$0.24 & 98.14$\pm$0.23 & \textbf{83.75$\pm$0.30} & 97.12$\pm$0.33 \\

SENet154~\cite{hu2018squeeze} & 89.80$\pm$0.37 & 97.92$\pm$0.27 & 81.96$\pm$0.37 & 96.90$\pm$0.25 \\

SENet154~\cite{hu2018squeeze} + PHG & 90.83$\pm$0.23 & 98.67$\pm$0.30 & 82.04$\pm$0.41& 97.12$\pm$0.19\\

SwinV2-B~\cite{liu2022swin} & 90.85$\pm$0.34 & 97.99$\pm$0.38 & 82.23$\pm$0.27 & \textbf{97.32$\pm$0.27} \\ 

SwinV2-B~\cite{liu2022swin} + PHG & \textbf{91.92$\pm$0.27} & \textbf{98.97$\pm$0.42}& 83.14$\pm$0.36 & 97.28$\pm$0.35\\ 
        
$p$-value & 0.0006 & 0.005 & 0.002 & -- \\ 
\end{tabular}	

\end{table*}

\begin{table*}[t!]
\centering

\caption{Experimental results on the Prostate Cancer dataset. 
Reported $p$-values reflect statistical significance of the improvement achieved by SwinV2-B + PHG over SwinV2-B, assessed by paired t-test. 
} 

\label{main_psc}
\begin{tabular}{l|c|c|c|c}

Method &  Accuracy & AUC & Sensitivity & Specificity \\

\hline
  
ResNet152~\cite{resnet} & 92.96$\pm$0.28 & 98.02$\pm$0.21 & 96.54$\pm$0.18 & 97.44$\pm$0.34 \\
  
ResNet152~\cite{resnet} + PHG & 94.01$\pm$0.28 & 98.80$\pm$0.30 & 96.91$\pm$0.22 & 96.67$\pm$0.41\\
  
SENet154~\cite{hu2018squeeze} & 93.73$\pm$0.39 & 98.67$\pm$0.34  & 94.84$\pm$0.50 & 96.38$\pm$0.28\\
  
SENet154~\cite{hu2018squeeze} + PHG & 97.99$\pm$0.21  & 99.72$\pm$0.29 & 96.55$\pm$0.33 & \textbf{98.26$\pm$0.40}\\

SwinV2-B~\cite{liu2022swin} & 95.21$\pm$0.33& 98.61$\pm$0.31 & 97.22$\pm$0.34 & 97.99$\pm$0.29\\
        
SwinV2-B~\cite{liu2022swin} + PHG & \textbf{98.64$\pm$0.27} & \textbf{99.83$\pm$0.24} & \textbf{98.34$\pm$0.29} & 98.17$\pm$0.17\\
        
$p$-value & $\ll 0.001$ & 0.0001 & 0.0005 & 0.265 \\
\end{tabular}	

\end{table*}

\begin{table*}[t!]
\centering

\caption{Experimental results on the CBIS-DDSM dataset
demonstrate that SwinV2-B + PHG significantly outperforms all the other tested methods for all the evaluation metrics.
The $p$-values are obtained by paired t-test between SwinV2-B + PHG and SwinV2-B.}

\label{main_cbis}

\begin{tabular}{l|c|c|c|c}

Method &  Accuracy & AUC & Sensitivity & Specificity \\

\hline

ResNet152~\cite{resnet} & 71.16$\pm$0.34 & 78.75$\pm$0.45 & 73.19$\pm$0.55 & 71.65$\pm$1.08 \\
		
ResNet152~\cite{resnet} + PHG & 74.34$\pm$0.31 & 79.35$\pm$0.36 & 71.74$\pm$1.10 & 72.03$\pm$0.49\\		

SENet154~\cite{hu2018squeeze} & 71.96$\pm$0.19 & 79.01$\pm$0.29 & 72.02$\pm$0.34 & 71.52$\pm$0.62\\

SENet154~\cite{hu2018squeeze} + PHG & 75.66$\pm$0.26 & 82.23$\pm$0.42 & 73.53$\pm$0.29 & 73.07$\pm$0.21\\

SwinV2-B~\cite{liu2022swin} & 73.51$\pm$0.22 & 81.58$\pm$0.54 &72.92$\pm$0.31 & 72.28$\pm$0.20 \\
        
SwinV2-B~\cite{liu2022swin} + PHG & \textbf{77.23$\pm$0.37} & \textbf{83.39$\pm$0.43}& \textbf{75.89$\pm$0.29} & \textbf{74.83$\pm$0.38} \\ 
        
$p$-value & $\ll 0.001$ & 0.0004  & $\ll 0.001$ & $\ll 0.001$\\
		
	\end{tabular}
\end{table*}

\setlength{\tabcolsep}{2pt}
\begin{table*}[t!]
\centering

\caption{Ablation study on the ISIC 2018 dataset. Reported $p$-$\text{value}_1$  was obtained using paired t-test between settings 12 and 13 in this table, measuring the statistical significance of the differences between KD and our PHG. ``--'' is reported since the Spe score of 13 is lower than 12. Reported $p$-$\text{value}_2$ was obtained using paired t-test between settings 10 and 13, measuring the statistical significance between PLLay and our PD encoder. ``Concat" means concatenating CNN features and topological features at the last CNN layer.
}

\label{ablation_isic}
\begin{tabular}{r|cc|ccc|ccc|cccc}
   
& \multicolumn{2}{c|}{Backbone} & \multicolumn{3}{c|}{PD Module}  & \multicolumn{3}{c|}{PD Fusion} &  \multirow{2}{*}{Acc} & \multirow{2}{*}{AUC} & \multirow{2}{*}{Sen} & \multirow{2}{*}{Spe} \\

\cline{2-9}  

& SENet154 & SwinV2-B & PersLay & PLLay & Our PD & Concat & KD & PHG & & & & \\
    
\hline

1 & & & $\checkmark$ & & & & & & 65.69$\pm$0.26 & 82.30$\pm$0.55 & 36.48$\pm$0.21 & 91.72$\pm$0.36 \\

2& & &  & $\checkmark$ & & & & & 67.13$\pm$0.26 & 82.02$\pm$0.49 & 32.12$\pm$0.25 & 91.97$\pm$0.48 \\

3 & & &  &  & $\checkmark$ & & & & 70.72$\pm$0.43 & 91.66$\pm$0.52 & 53.01$\pm$0.72 & 94.62$\pm$0.36 \\

     \hline

4 & $\checkmark$ & & $\checkmark$ &  & & & & $\checkmark$& 90.02$\pm$0.68 & 97.37$\pm$0.38 & 81.80$\pm$0.61 & 97.01$\pm$0.70 \\

5 & $\checkmark$ &  &  & $\checkmark$ & & & & $\checkmark$ & 90.27$\pm$0.21 & 97.04$\pm$0.65 & 80.66$\pm$0.71 & 96.82$\pm$0.45 \\

    
6 & $\checkmark$ &  &  & & $\checkmark$ & $\checkmark$ & &  & 89.97$\pm$0.34 & 97.34$\pm$0.47 & 82.21$\pm$0.57&96.50$\pm$0.39\\

7 & $\checkmark$ &  &  & & $\checkmark$ &  & $\checkmark$ &  & 90.27$\pm$0.29 & 96.94$\pm$0.48 & 82.09$\pm$0.37 & 96.47$\pm$0.64\\

8 & $\checkmark$ &  &  & & $\checkmark$ & & & $\checkmark$ & 90.83$\pm$0.23 & 98.67$\pm$0.30 & 82.04$\pm$0.41& 97.12$\pm$0.19\\

\hline
    
9 & & $\checkmark$ & $\checkmark$ & & &  & & $\checkmark$ & 91.00$\pm$0.44 & 98.23$\pm$0.17 & 81.76$\pm$0.19 & 97.40$\pm$0.25\\
     
10 & & $\checkmark$ & & $\checkmark$ & & & & $\checkmark$ & 91.12$\pm$0.29& 98.46$\pm$0.31 & 82.33$\pm$0.52 & 96.78$\pm$0.41 \\
     
11 & & $\checkmark$ & & & $\checkmark$ & $\checkmark$ &  &  & 91.18$\pm$0.19 & 98.09$\pm$0.26 & 81.99$\pm$0.24 & \textbf{97.68$\pm$0.37}\\
     
12 & & $\checkmark$ & & & $\checkmark$ & &$\checkmark$ & & 91.00$\pm$0.44 & 98.20$\pm$0.38 & 82.42$\pm$0.59&97.45$\pm$0.40\\
     
13 & & $\checkmark$  & & & $\checkmark$ & & & $\checkmark$ & \textbf{91.92$\pm$0.27} & \textbf{98.97$\pm$0.42}& \textbf{83.14$\pm$0.36} & 97.28$\pm$0.35 \\

\hline
     
\multicolumn{9}{c|}{$p$-$\text{value}_1$} & 0.016 & 0.004 & 0.048 & -- \\


\multicolumn{9}{c|}{$p$-$\text{value}_2$} & 0.002 & 0.06 & 0.021 & 0.072 \\
     
\end{tabular}

\end{table*}
\setlength{\tabcolsep}{2pt}
\begin{table*}[t!]
\centering

\caption{Ablation study on the Prostate Cancer dataset. Reported $p$-$\text{value}_1$  was obtained using paired t-test between settings 12 and 13 in this table, measuring the statistical significance between KD and our PHG. Reported $p$-$\text{value}_2$ was obtained using paired t-test between settings 10 and 13, measuring the statistical significance between PLLay and our PD encoder. ``Concat" means concatenating CNN features and topological features at the last CNN layer.}

\label{ablation_psc}

\begin{tabular}{r|cc|ccc|ccc|cccc}
   & \multicolumn{2}{c|}{Backbone} & \multicolumn{3}{c|}{PD Module}  & \multicolumn{3}{c|}{PD Fusion} &  \multirow{2}{*}{Acc} & \multirow{2}{*}{AUC} & \multirow{2}{*}{Sen} & \multirow{2}{*}{Spe} \\

\cline{2-9}  

& SENet154 & SwinV2-B & PersLay & PLLay & Our PD & Concat & KD & PHG & & & & \\
    
\hline

1 & & & $\checkmark$ & & & & & & 75.31$\pm$0.46 & 86.43$\pm$0.53 &67.37$\pm$0.57 &84.42$\pm$0.38 \\

2 & & &  & $\checkmark$ & & & & & 77.34$\pm$0.30 & 89.84$\pm$0.56& 66.58$\pm$0.33 &85.75$\pm$0.22 \\

3 & & &  &  & $\checkmark$ & & & & 82.35$\pm$0.46 & 92.53$\pm$0.20 & 76.37$\pm$0.48 & 87.46$\pm$0.58 \\

\hline

4 & $\checkmark$ & & $\checkmark$ &  & & & & $\checkmark$& 96.91$\pm$0.43 & 99.68$\pm$0.25 & 95.92$\pm$0.40 &98.09$\pm$0.44 \\

5 & $\checkmark$ &  &  & $\checkmark$ & & & & $\checkmark$ & 97.11$\pm$0.30 & 99.68$\pm$0.43 & 96.13$\pm$0.47 & 98.21$\pm$0.39 \\

    
6 & $\checkmark$ &  &  & & $\checkmark$ & $\checkmark$ &  &  & 97.02$\pm$0.55 & 99.45$\pm$0.39&96.28$\pm$0.30& \textbf{98.40$\pm$0.47}\\

7 & $\checkmark$ &  &  & & $\checkmark$ &  & $\checkmark$ &  & 96.57$\pm$0.29 & 99.55$\pm$0.27&96.57$\pm$0.29 & 97.91$\pm$0.28\\

8 & $\checkmark$ &  &  & & $\checkmark$ & & & $\checkmark$ & 97.99$\pm$0.21  & 99.72$\pm$0.29 & 96.55$\pm$0.33 & 98.26$\pm$0.40\\
     
\hline

9 & & $\checkmark$ & $\checkmark$ & & &  & & $\checkmark$ & 98.21$\pm$0.31&99.54$\pm$0.46 & 97.54$\pm$0.38&97.59$\pm$0.24\\
    
10 & & $\checkmark$ & & $\checkmark$ & & & & $\checkmark$ &98.04$\pm$0.26 &99.06$\pm$0.27 &97.31$\pm$0.24&97.54$\pm$0.19 \\
     
11 & & $\checkmark$ & & & $\checkmark$ & $\checkmark$ & & &98.04$\pm$0.29 &98.12$\pm$0.25 &\textbf{98.45$\pm$0.41} & 97.12$\pm$0.28\\
     
12 & & $\checkmark$ & & & $\checkmark$ & &$\checkmark$ & & 98.11$\pm$0.40&99.75$\pm$0.21 & 97.24$\pm$0.32&97.77$\pm$0.22 \\
     
13 & & $\checkmark$   & & & $\checkmark$ & & & $\checkmark$  &   \textbf{98.64$\pm$0.27} & \textbf{99.83$\pm$0.24} & 98.34$\pm$0.29 & 98.17$\pm$0.17 \\
    
\hline
    
\multicolumn{9}{c|}{$p$-$\text{value}_1$} & 0.039 & 0.59 & 0.0004 & 0.012\\

\multicolumn{9}{c|}{$p$-$\text{value}_2$} & 0.007& 0.001& 0.0002 & $\ll$0.001
\end{tabular}

\end{table*}
\setlength{\tabcolsep}{2pt}
\begin{table*}[t!]
\centering

\caption{Ablation study on the CBIS-DDSM dataset. Reported $p$-$\text{value}_1$  was obtained using paired t-test between settings 12 and 13 in this table, measuring the statistical significance between KD and our PHG. Reported $p$-$\text{value}_2$ was obtained using paired t-test between settings 10 and 13, measuring the statistical significance between PLLay and our PD encoder. ``Concat" means concatenating CNN features and topological features at the last CNN layer.}

\label{ablation_cbis}

\begin{tabular}{r|cc|ccc|ccc|cccc}
    & \multicolumn{2}{c|}{Backbone} & \multicolumn{3}{c|}{PD Module}  & \multicolumn{3}{c|}{PD Fusion} &  \multirow{2}{*}{Acc} & \multirow{2}{*}{AUC} & \multirow{2}{*}{Sen} & \multirow{2}{*}{Spe} \\

\cline{2-9}  

& SENet154 & SwinV2-B & PersLay & PLLay & Our PD & Concat & KD & PHG & & & & \\
    
\hline

1 & & & $\checkmark$ & & & & & & 61.38$\pm$0.32 & 55.05$\pm$0.10 & 55.60$\pm$0.25 &57.72$\pm$0.14 \\

2 & & &  & $\checkmark$ & & & & & 63.23$\pm$0.26 & 63.06$\pm$0.21 &60.36$\pm$0.37 &59.40$\pm$0.35 \\

3 & & &  &  & $\checkmark$ & & & & 64.02$\pm$0.31 & 62.44$\pm$0.48 & 58.29$\pm$0.44 & 60.17$\pm$0.19 \\

     \hline

4 & $\checkmark$ & & $\checkmark$ &  & & & & $\checkmark$& 74.87$\pm$0.30 & 79.71$\pm$0.33& 74.00$\pm$0.35 &73.13$\pm$1.31 \\

5 & $\checkmark$ &  &  & $\checkmark$ & & & & $\checkmark$ & 74.60$\pm$0.40 & 81.66$\pm$0.32 &75.02$\pm$0.25 &73.69$\pm$0.35 \\

    
6 & $\checkmark$ &  &  & & $\checkmark$ & $\checkmark$ & & &  72.36$\pm$0.64& 80.27$\pm$0.57 &73.58$\pm$0.47 & 72.89$\pm$0.46 \\

7 & $\checkmark$ &  &  & & $\checkmark$ &  & $\checkmark$ &  & 73.01$\pm$0.47 & 81.90$\pm$0.35&74.40$\pm$0.46&73.01$\pm$0.77\\

8 & $\checkmark$ &  &  & & $\checkmark$ & & & $\checkmark$ & 75.66$\pm$0.26 & 82.23$\pm$0.42 & 73.53$\pm$0.29 & 73.07$\pm$0.21\\

\hline

9 & & $\checkmark$ & $\checkmark$ & & &  & & $\checkmark$ & 73.71$\pm$0.30 &81.26$\pm$0.27 & 72.64$\pm$0.29&72.54$\pm$0.30\\
     
10 & & $\checkmark$ & & $\checkmark$ & & & & $\checkmark$ & 74.06$\pm$0.45&82.21$\pm$0.53 &73.45$\pm$0.47&73.04$\pm$0.42 \\
     
11 & & $\checkmark$ & & & $\checkmark$ & $\checkmark$ & & & 73.58$\pm$0.27&82.04$\pm$0.51 &74.11$\pm$0.45 &73.44$\pm$0.51\\
     
12 & & $\checkmark$ & & & $\checkmark$ & &$\checkmark$ & & 74.34$\pm$0.40&82.41$\pm$0.43&74.03$\pm$0.46&73.87$\pm$0.29 \\
     
13 & & $\checkmark$  & & & $\checkmark$ & & & $\checkmark$  & \textbf{77.23$\pm$0.37} & \textbf{83.39$\pm$0.43}& \textbf{75.89$\pm$0.29} & \textbf{74.83$\pm$0.38} \\

\hline

\multicolumn{9}{c|}{$p$-$\text{value}_1$} &$\ll$0.001 & 0.007 & $\ll$0.001 & 0.002\\

\multicolumn{9}{c|}{$p$-$\text{value}_1$} & $\ll$0.001 & 0.005& $\ll$0.001 & 0.0001
     
\end{tabular}

\end{table*}



Common data augmentation techniques such as random flipping, color jittering, random rotation, and cropping are used. The network is optimized using the Adam optimizer with an initial learning rate of 0.0001, $\beta_1$ value of 0.9, and $\beta_2$ value of 0.999. A polynomial learning rate decay with a power of 0.9 is applied. The maximum number of training epochs is set to 1000. The value of $\alpha$ in Eq.~(\ref{eq-final-loss}) is set to 0.1. The parameter $r$ for balancing the trade-off between performance and complexity is empirically set to 8. All experiments are conducted on an NVIDIA P100 GPU using PyTorch. Four common evaluation metrics are used: accuracy (Acc), area under the receiver operating characteristic (ROC) curve (AUC), sensitivity (Sen), and specificity (Spe).

We utilize the GUDHI package~\cite{gudhi} to compute persistence diagrams. The points in a persistence diagram are first scaled to the range of $[0, 1]$ and then normalized. Additionally, a one-hot marker indicating which homology group each point belongs to is added. Any persistence value smaller than 10 is ignored. The points within each homology group are sorted in decreasing order based on their persistence values. For each persistence diagram, we experimented with selecting the top $n$ = 50, 100, 150, 200, and 250 points (see Eq.~(\ref{eq-PD-encoding})). We found that selecting $n$ = 150 points yields saturated performance while maintaining relatively low computation, as most of the points are very close to the diagonal line and are considered as noise points.

\vspace{-0.0cm}
Moreover, we add dummy points of $(0, 0)$ to a persistence diagram if the number of points in it is less than 150. All the points are given an additional coordinate to indicate that a point is an added one (marked by 0) or originally presented one (marked by 1).
\setlength{\tabcolsep}{3pt}
\begin{table*}[t!]
\centering

\caption{Time and parameter complexities of our PHG-Net approach on an NVIDIA P100 GPU; w/ share and w/o share represent using one PD encoder and multiple PD encoders for multi-scale vision feature fusion, respectively.}\label{para_flops}
\centering
\begin{tabular}{l|l|c|c|c|c|c|c}
\multirow{2}{*}{Method} & \multirow{2}{*}{Input Size} & \multicolumn{3}{c|}{Acc} & \multirow{2}{*}{\# Params.} & \multirow{2}{*}{FLOPs} & \multirow{2}{*}{FPS} \\
& & ISIC 2018 & Prostate & CBIS-DDSM & & & \\
\hline

SwinV2-B & (3, 224, 224) & 90.85 & 95.21&73.51 & 86.913 M & 20.370 G & 14.937 \\

SwinV2-B+PLLay+PHG & (3, 224, 224) (4, 300) & 91.12 & 98.04 & 74.06 & 95.073 M & 20.375 G & 20.787\\

SwinV2-B+PD+PHG w/ share & (3, 224, 224) (4, 300) & 91.92& 98.64& 77.23& 98.065 M & 20.796 G & 15.914 \\

SwinV2-B+PD+PHG w/o share & (3, 224, 224) (4, 300) & 91.95 & 98.70& 77.04& 108.892 M & 22.062 G & 17.059\\
            
\end{tabular}

\end{table*}
\subsection{Results and Analysis}

We evaluate the effectiveness and robustness of our PHG-Net approach by utilizing two common CNN backbones (ResNet152~\cite{resnet} and SENet154~\cite{hu2018squeeze}), which performed well on medical image datasets (e.g., used by the team attaining top-1 score in the ISIC 2018 Challenge), as well as a latest Transformer, Swin Transformer v2 (SwinV2-B)~\cite{liu2022swin}. These models are evaluated without using extra training data. To demonstrate the statistical significance of the performance improvements yielded by our PHG-Net, we compute $p$-values (with paired $t$-test) between SwinV2-B + PHG (our model) and SwinV2-B by running each experiment 5 rounds. For each round, we select 5 epochs that achieve the top-5 best accuracy and take their average as the final result for robust evaluation.

The ISIC 2018 and Prostate Cancer datasets are for multi-class classification tasks. Thus, we report the OvR (One-vs-Rest) scores for AUC, Specificity, and Sensitivity. In each time, we take one class as positive and the other classes as negative, under which we compute the score for each class $c$. Finally, the average score of all the classes is reported.

Tables~\ref{main_isic}, \ref{main_psc}, and \ref{main_cbis} present the experimental results. We find that our PHG-Net approach improves, for example, the accuracy by 1.07\%, 3.43\%, and 3.72\% on the ISIC 2018, prostate cancer, and CBIS-DDSM datasets, respectively, when using the SwinV2-B backbone. These performance improvements and the $p$-values verify the effectiveness of our approach, especially on the prostate cancer and breast mammography datasets, where discernible changes in appearances or topological structures occur when cancer is present. For instance, in an H\&E-stained prostate image, the number of nuclei is likely to increase if cancer cells are found in the image. In breast mammography images, tumor areas are often characterized by dense masses or clusters of micro-calcifications. In some cases, they may be present as tiny calcium deposits that appear as small white specks or clusters in the images. These topological structures, such as connected components or loops, can differentiate the images containing them from normal images.

\begin{figure*}[!t]

\centering
\includegraphics[width=1.2\columnwidth]{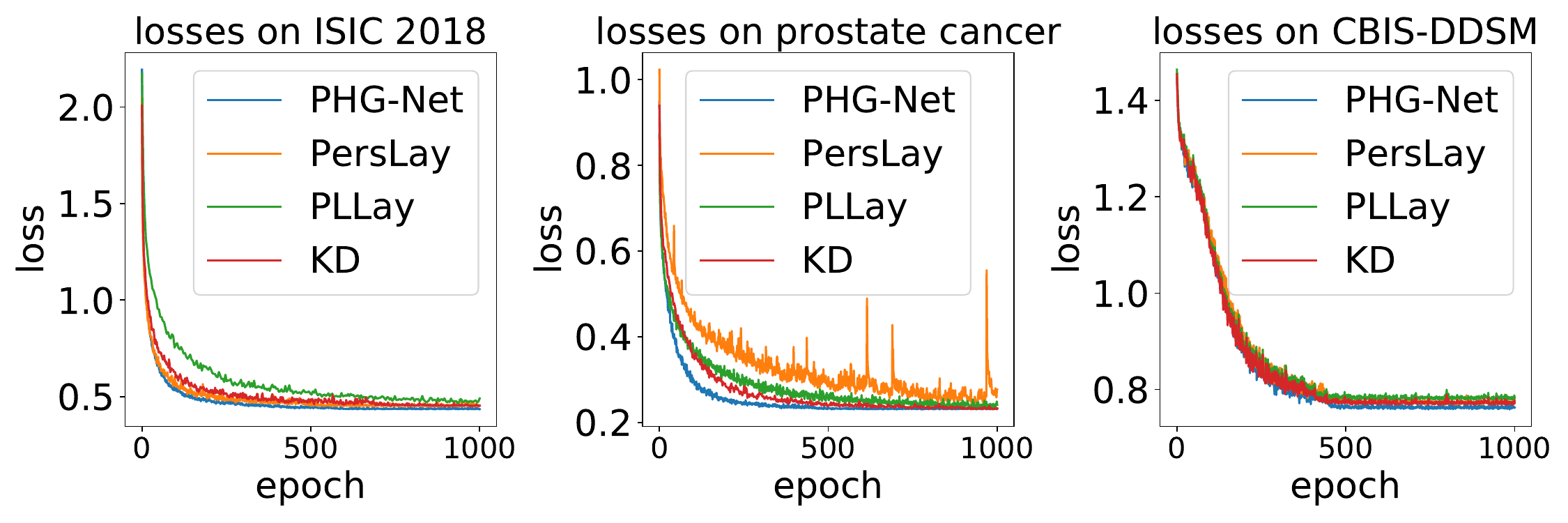}

\caption{Validation losses of different persistent homology based methods and the KD (with our PD encoder) method.}
\label{loss}
\end{figure*}

\subsection{Ablation Study}
\begin{figure*}[ht!]

    \centering
    \includegraphics[width=1.12\columnwidth]{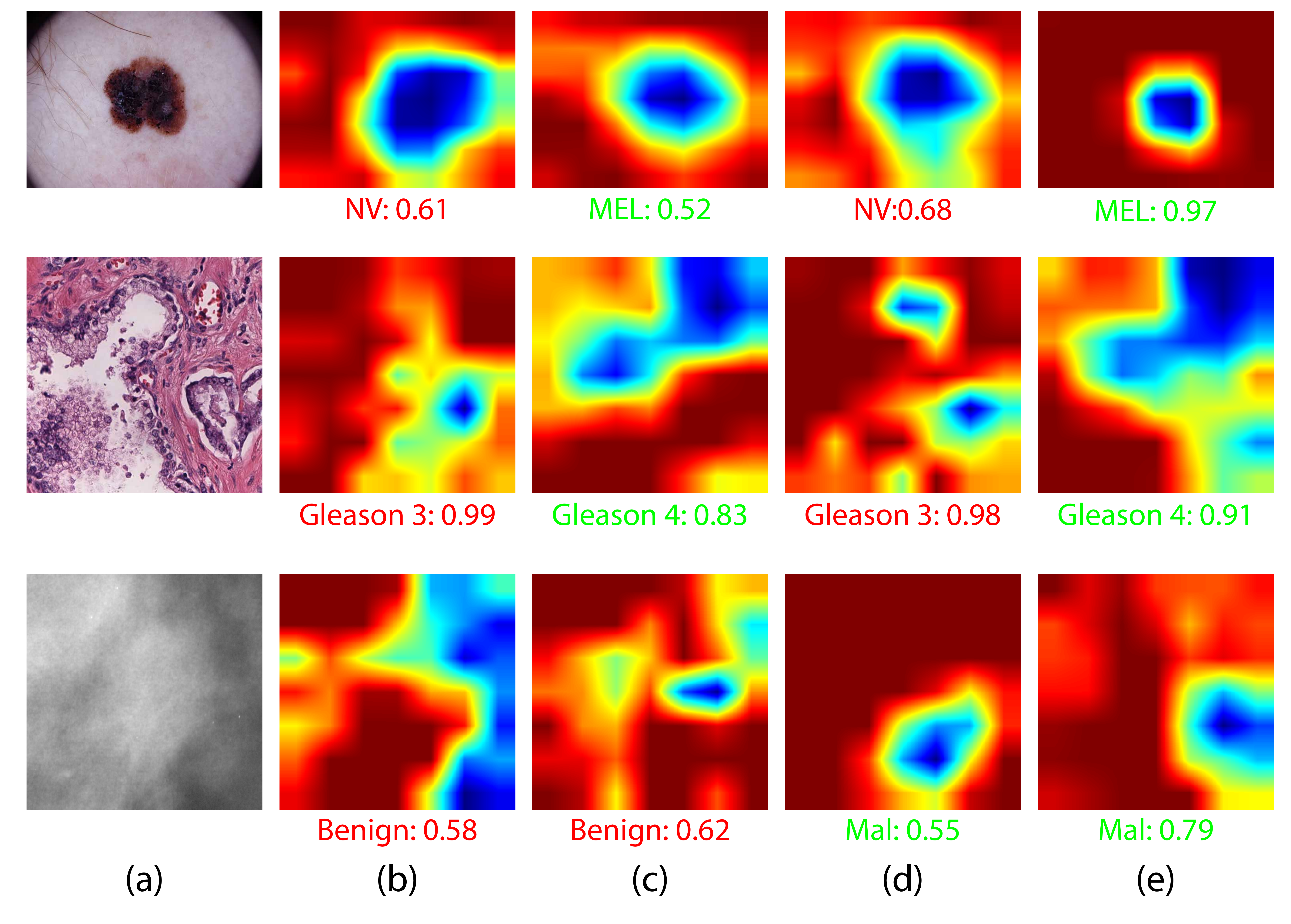}
    \caption{Qualitative examples of different persistent homology based methods and the KD (with our PD encoder) method. The 1st, 2nd, and 3rd rows are for samples from the ISIC 2018, Prostate Cancer, and CBIS-DDSM datasets, respectively. Columns (a)-(e) are for input images, heat-maps by PLLay, PersLay, KD, and PHG-Net, respectively. \textcolor{green}{{\bf Green}} and \textcolor{red}{{\bf red}} correspond to correct and incorrect predictions, respectively. The \textcolor{green}{green} or \textcolor{red}{red} floating-point number under each heat-map represents the top-1 prediction confidence (the higher the better).}
    \label{result}
\end{figure*}
To examine the effects of our new persistence diagram (PD) encoder and PHG mechanism, we compare our methods with two common persistence diagram vectorization-based methods: (1) PersLay~\cite{perslay} (it encodes a persistence diagram and then refines the PD using an FC layer, where the persistence silhouette setting~\cite{chazal2014stochastic} is applied), and (2) PLLay~\cite{pllay} (it utilizes a weighted persistent landscape to extract topological features). Specifically, we pass the topological features extracted by these methods through the PH branch and compare their performances. Furthermore, we investigate the effect of using only topological features in the models. Additionally, we compare our approach with a typical knowledge distillation method (KD) \cite{du2022distilling}, but replace its backbone with SENet154 and SwinV2-B, respectively, for fair comparison.

Tables~\ref{ablation_isic}, \ref{ablation_psc}, and \ref{ablation_cbis} show that our PD encoder outperforms the PersLay and PLLay methods, while our PHG outperforms the KD and concatenation (Concate, which concatenates CNN features and topological features at the last CNN layer) methods, since our PHG refines CNN features at multiple scales. This confirms the effectiveness of our schemes for constructing the PD encoder and PH guidance. It is worth noting that the methods relying {\bf solely} on topological information (PersLay, PLLay, and our PD encoder) do not perform well on these datasets. 
Note that PersLay is a general form of persistence landscape~\cite{pers_land}, persistence silhouette~\cite{chazal2014stochastic}, and persistence images~\cite{pers_image} by utilizing different hyper-parameters, and we found that there is no obvious difference among the settings of these PH methods. Thus, our approach also outperforms these PH methods. The $p$-values in Tables~\ref{ablation_isic}, \ref{ablation_psc}, and \ref{ablation_cbis} demonstrate the statistical significance of our PD encoder and PHG module.

\subsection{Time and Parameter Complexity}

To examine the time and parameter complexities of our PHG-Net, we present the number of parameters, floating point operations (FLOPs), and frames per second (FPS) for our approach and the SwinV2-B model in Table~\ref{para_flops} for comparison.

In Table~\ref{para_flops}, $(3, 224, 224)$ denotes the size of an input image, while $(4, 300)$ stands for the size of the corresponding persistence diagram. It can be observed from Table~\ref{para_flops} that the additional parameters and computation costs caused by our PHG-Net are quite limited. Furthermore, sharing the PD Encoder at multiple scales will result in fewer parameters and FLOPs without reducing the performance.





\subsection{Qualitative Results}
The validation losses of different persistent homology based methods and the KD method on the three datasets using the backbone of SwinV2-B \cite{hu2018squeeze} are shown in Fig.~\ref{loss}.

Some qualitative examples based on the backbone of SwinV2-B on the three datasets are shown in Fig.~\ref{result}, which demonstrate that our PHG-Net is capable of helping the network concentrate on the correct regions.

\section{Conclusions}
In this paper, we proposed to extract and encode persistence diagrams of medical images as important topological features for classification tasks using a neural network (PD encoder) that we designed. Our new approach, PHG-Net, is data-driven, learnable, and superior in performance over known topology based methods. We further developed a persistence diagram (PD) guided mechanism, which incorporates PD features into CNN or Transformer for co-training of DL models. Experiments on three datasets validated the effectiveness of our PHG-Net. Complexity analysis showed that the extra costs introduced by our approach are quite small. It is expected that our PHG-Net approach will provide a new perspective and paradigm for combining topological features with CNNs/Transformers to improve AI model performances for medical image analysis tasks.




{\small
\bibliographystyle{ieee_fullname}
\bibliography{egbib}
}

\end{document}